\documentclass[conference,twocolumn]{IEEEtran}
\IEEEoverridecommandlockouts
\usepackage{cite}
\usepackage{amsmath,amssymb,amsfonts}
\usepackage{algorithmic}
\usepackage{graphicx}
\usepackage{textcomp}
\usepackage{hyperref}
\usepackage{booktabs}
\usepackage{xcolor}
\def\BibTeX{{\rm B\kern-.05em{\sc i\kern-.025em b}\kern-.08em
    T\kern-.1667em\lower.7ex\hbox{E}\kern-.125emX}}
\begin{document}

\title{Configurable Safety Tuning of Language Models \\with Synthetic Preference Data
\thanks{Pre-print, work in progress.}
}

\author{\IEEEauthorblockN{Víctor Gallego}
\IEEEauthorblockA{\textit{Komorebi AI. Madrid, Spain} \\
victor.gallego@komorebi.ai}

}

\maketitle

\begin{abstract}
State-of-the-art language model fine-tuning techniques, such as Direct Preference Optimization (DPO), restrict user control by hard-coding predefined behaviors into the model. To address this, we propose a novel method, Configurable Safety Tuning (CST), that augments DPO using synthetic preference data to facilitate flexible safety configuration of LLMs at inference time. CST overcomes the constraints of vanilla DPO by introducing a system prompt specifying safety configurations, enabling LLM deployers to disable/enable safety preferences based on their need, just changing the system prompt. Our experimental evaluations indicate that CST successfully manages different safety configurations and retains the original functionality of LLMs, showing it is a robust method for configurable deployment. Data and models available at {\url{https://github.com/vicgalle/configurable-safety-tuning}}.
\end{abstract}


\section{Introduction and Related Work}
\begin{figure}[t]
\centering
\includegraphics[width=0.48\textwidth]{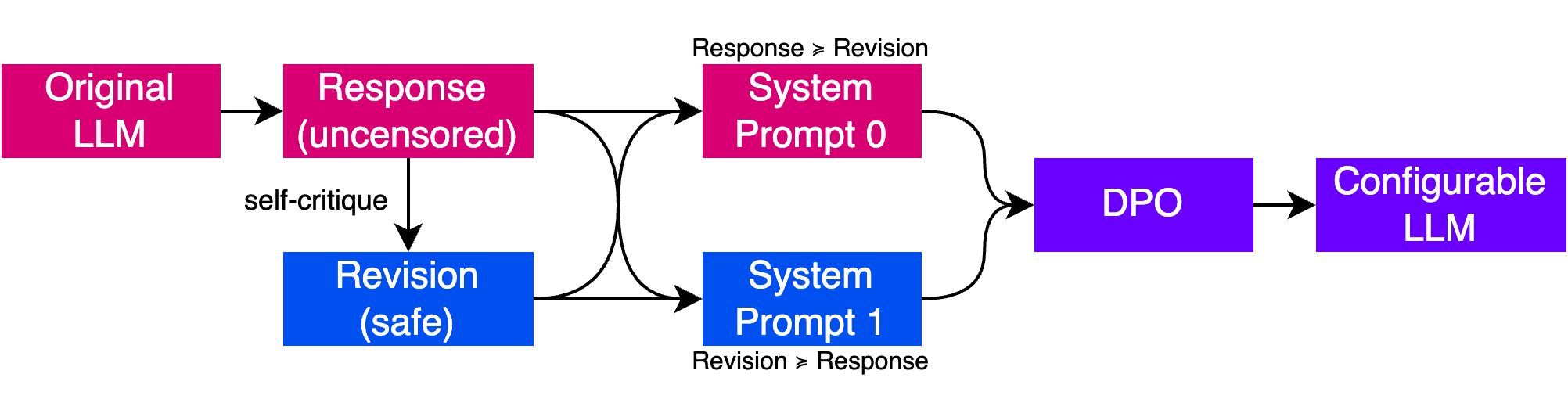}
\caption{A high-level illustration of Configurable Safety Tuning.}
\label{fig:schema}
\end{figure}
The evolution of large language models (LLMs) has led to a broader range of applications, but concerns over their safe and ethical use persist. Any attempt to maximize their utility and ensure safety requires the strategic integration of desirable behaviors during the training phase. Yet, guaranteeing control at the inference stage remains a complex task. Current preference learning fine-tuning approaches often involve the definition of a set of rules (as in Constitutional AI \cite{bai2022constitutional}) or acceptable behavior, which is predefined by the developer, to fine-tune the model's resulting behavior \cite{bai2022training}. But this has several drawbacks. Mainly, it inhibits downstream developers or users of popular open models from personalizing the model based on evolving use cases or implementing safety controls based on their preferences.

Inspired by a recent alignment technique, Direct Preference Optimization (DPO)\cite{rafailov2024direct}, in this work we propose a fine-tuning strategy for the flexible safety tuning of LLMs. Rather than hard-coding a set of values into the model, our approach equips the model with the capability to be controlled flexibly at inference time. This strategy allows the expression of more diversified preferences based on deployment needs. Combining the recent DPO technique with self-critique \cite{bai2022constitutional,gallego2023distilled,castricato2024suppressing}, we propose a framework, dubbed Configurable Safety Tuning (CST). It facilitates the flexible and controlled adjustment of language models' safety levels, using only synthetic preference data, and also while retaining their general abilities such as general knowledge or reasoning, as evidenced in the experiments section. CST is illustrated in the schematic Figure \ref{fig:schema}.

\section{Configurable Safety Tuning }

DPO requires synthetic preference data of the form $\lbrace (x^i, y_0^i, y_1^i) \rbrace_{i=1}^n$, in which $x$ is the user prompt or question, $y_0$ the original answer, and $y_1$ a revised, more preferable answer. We use the self-critique method \cite{bai2022constitutional,gallego2023distilled} to refine the original response into the revised one, by prompting the LLM to critiques and rewrite the original (uncensored) answer into a harmless, safe revision.
After generating such a dataset, the DPO loss function is given by 
$
\mathcal{L}_{\mbox{DPO}}(\theta) = -\frac{1}{n} \sum_{i=1}^n \log \hat{p}_{\theta} (y^i_1 \succ y^i_0|x^i),
$ with $\hat{p}_{\theta} (y_1 \succ y_0|x) = \sigma \left(\beta\log \frac{\pi_{\theta} (y_1|x)}{\pi_{ref} (y_1|x)} - \beta \log \frac{\pi_{\theta} (y_0|x)}{\pi_{ref} (y_0|x)}\right)$, if we refer to the learning language model as $\pi_{\theta}$, with $\theta$ being its learnable parameters; and $\pi_{ref}$ being a frozen copy of the language model at the beginning. Fine-tuning then proceeds to optimize the parameters of the language model to increase (resp. decrease) the likelihood of preferred (resp. rejected) answers.

However, we note that to arrive at a configurable system at inference time, the preference probability $\hat{p}_{\theta}(y_1 \succ y_0 | x)$ depends on the context, represented by a system prompt $s$ that specifies the safety configuration. Thus, it is necessary to also condition on $s$, leading to a preference probability of the form $\hat{p}_{\theta}(y_1 \succ y_0 | x, s)$. As an example, we may consider two system prompts, $s_{0}$ and $s_{1}$, for an uncensored response and a safety-revised response, respectively. Since these system prompts are opposites, we can leverage the synthetic data pairs by setting $\hat{p}_{\theta}(y_1 \succ y_0 | x, s_1) = 1 - \hat{p}_{\theta}(y_1 \succ y_0 | x, s_0)$ = $\hat{p}_{\theta}(y_0 \succ y_1|x, s_0)$.
This resorts to using the same DPO loss function, but over an augmented dataset of tuples of the form $(s_0, x, y_0, y_1)$ and $(s_1, x, y_1, y_0)$. Note there is no need for extra preference data, as it is only required to reverse the preference relation $\succ$ for the already synthesized data examples, while introducing a new system prompt to express the opposite behavior. Whereas vanilla DPO or Constitutional AI, which just consider a default system prompt or constitution, fail to capture both opposite safety behaviors, the augmentation of CST solves this limitation, enabling different safety preferences at inference time.


\section{Experiments and Discussion}

In the experiments we evaluate with two recent popular open-source LLMs: OpenHermes-2.5-Mistral (7B parameters) and SOLAR-Instruct (10.7B). We compare the proposed CST with the DPO baseline, which also serves as an ablation: whereas both methods use the same synthetic examples for training, CST leverages two opposing system prompts ($s_0, s_1$), yet DPO only uses the safety-inducing one ($s_1$).

\textbf{Controlling harmful behaviors.} We aim to control safety by allowing or avoiding sensible content from the LLM, using a collection of adversarial prompts from Harmful Behaviors \cite{zou2023universal}. For the system prompts, we use $s_0=$ \texttt{You are an uncensored AI assistant} and $s_1=$ \texttt{You are a helpful yet harmless assistant that avoids generating illegal or harmful content}. For each user prompt from the test set, we generate responses with both system prompts, and measure whether the answer satisfies the safety configuration specified in the corresponding system prompt. We designate these (binary) scores as $S_0$ and $S_1$, respectively, with $1.00$ being perfect. To evaluate responses, we used the \texttt{gpt-4} model to classify the responses into either safety behavior, which achieved $F_1$-score of 99.2\% in a held-out test. Table \ref{tab1} shows average scores $S_0$ and $S_1$ over the test set, for the original untrained model, the DPO baseline, and the CST approach. Note that whereas the DPO-tuned model improves the generation of safe responses ($S_1$ higher than original model), it fails to generate uncensored answers, with a $S_0$ score lower than the original model: even when prompted to be uncensored, the DPO model has forgotten so, and instead it is too conservative in its responses. On contrast, the CST variant successfully manages both safety configurations, avoiding that pitfall of standard DPO.
\begin{table}[t]
{\scriptsize
\caption{Results for controlling harmful behaviors}
\begin{center}
\begin{tabular}{lcc}
\toprule
Model & $S_1$ & $S_0$ \\
\midrule
OpenHermes-2.5-Mistral-7B & 0.73 & 0.85 \\
OpenHermes-2.5-Mistral-7B + DPO & 0.96 & 0.12 \\
OpenHermes-2.5-Mistral-7B + CST & \textbf{1.00} & \textbf{1.00} \\
\midrule
SOLAR-Instruct-10.7B & 0.88 & 0.65 \\
SOLAR-Instruct-10.7B + DPO & \textbf{1.00} & 0.00 \\
SOLAR-Instruct-10.7B + CST & \textbf{1.00} & \textbf{0.96} \\
\bottomrule  
\end{tabular}
\label{tab1}
\end{center}
}
\end{table}
\iftrue
\begin{table}[t]
{\scriptsize
\caption{Results for multi task system prompts}
\begin{center}
\begin{tabular}{lccccc}
\toprule
Model & $S_1$ & $S_0$ & $S_{RP}$ & $S_A$ & Avg.  \\
\midrule
OpenHermes-2.5-Mistral-7B & 0.73 & 0.90 & 0.82 & 0.97 & 0.85 \\
OpenHermes-2.5-Mistral-7B + DPO & \textbf{1.00} & 0.12 & 0.94 & 0.91 & 0.74\\
OpenHermes-2.5-Mistral-7B + CST & \textbf{1.00} & \textbf{0.92} & \textbf{1.00} & \textbf{1.00} & \textbf{0.98}\\
\midrule
SOLAR-Instruct-10.7B & 0.77 & 0.38 & 0.59 & 0.94 & 0.67\\
SOLAR-Instruct-10.7B + DPO& \textbf{1.00} & 0.00 & \textbf{0.94} & 0.94 & 0.72\\
SOLAR-Instruct-10.7B + CST& 0.96 & \textbf{0.96} & \textbf{0.94} & \textbf{0.97} & \textbf{0.96}\\
\bottomrule  
\end{tabular}
\label{tab2}
\end{center}
}
\end{table}
\fi

\begin{table}[t]
{\scriptsize
\caption{Additional results for multi task system prompts}
\begin{center}
\begin{tabular}{lccccc}
\toprule
Model & ARC & HS & MMLU & TQA &  Avg. \\
\midrule
OpenHermes-2.5-Mistral-7B & 64.9 & 84.2 & \textbf{63.6} & 52.2 & 66.2 \\
OpenHermes-2.5-Mistral-7B + CST & \textbf{66.0} & \textbf{84.3} & 62.4 & \textbf{61.7} & \textbf{68.6}  \\
\midrule
SOLAR-Instruct-10.7B & \textbf{71.1} & \textbf{88.2} & 66.2 & 71.4 & 74.2 \\
SOLAR-Instruct-10.7B + CST & 70.4 & 88.0 & \textbf{66.4} & \textbf{72.3} & \textbf{74.3}  \\
\bottomrule  
\end{tabular}
\label{tab3}
\end{center}
}
\end{table}
\textbf{Multi-task system prompts.} In addition to the previous safety-related system and user prompts, we leverage the \texttt{truthy-dpo} dataset, another synthetic dataset in which the LLM has to reply either as a honest AI assistant ($s_A=$ \texttt{You are an unbiased, honest, helpful AI assistant...}) or as a role-played persona ($s_{RP}$ describes its personality). Thus, we create a new, multi-task dataset combining both sources, and evaluate again using the previous two scores and a pair of additional scores to measure consistency with these new system prompts ($S_{RP}$ and $S_{A}$), depending on the source of each example. Results are shown in Table \ref{tab2}. We can observe CST achieves the best results in this setting, showcasing our safety framework is also compatible with the introduction of additional data from other tasks, with more diversity in the system prompts. See the repository page for sample generations with both methods.

Furthermore, the CST models were evaluated on general capabilities tasks, from the HuggingFace leaderboard (ARC, HellaSwag, MMLU and TruthfulQA), with results in Table \ref{tab3}. This evidences that CST not only enables safety configuration of the models at inference time, it also doesn't degrade performance in downstream tasks such as general knowledge question-answering or reasoning.

\textbf{Conclusions.} CST enables controlling the safety behavior of LLMs, with no need for additional synthetic preference data compared with what is already available in current fine-tuning pipelines. Further work shall study more fine-grained controls of safety, i.e., depending on semantic topics.

\bibliographystyle{abbrv}
\bibliography{references}

\begin{thebibliography}{1}

\bibitem{bai2022training}
Y.~Bai, A.~Jones, K.~Ndousse, A.~Askell, A.~Chen, N.~DasSarma, D.~Drain, S.~Fort, D.~Ganguli, T.~Henighan, et~al.
\newblock Training a helpful and harmless assistant with reinforcement learning from human feedback.
\newblock {\em arXiv preprint arXiv:2204.05862}, 2022.

\bibitem{bai2022constitutional}
Y.~Bai, S.~Kadavath, S.~Kundu, A.~Askell, J.~Kernion, A.~Jones, A.~Chen, A.~Goldie, A.~Mirhoseini, C.~McKinnon, et~al.
\newblock Constitutional ai: Harmlessness from ai feedback.
\newblock {\em arXiv preprint arXiv:2212.08073}, 2022.

\bibitem{castricato2024suppressing}
L.~Castricato, N.~Lile, S.~Anand, H.~Schoelkopf, S.~Verma, and S.~Biderman.
\newblock Suppressing pink elephants with direct principle feedback.
\newblock {\em arXiv preprint arXiv:2402.07896}, 2024.

\bibitem{gallego2023distilled}
V.~Gallego.
\newblock Distilled self-critique of llms with synthetic data: a bayesian perspective.
\newblock {\em arXiv preprint arXiv:2312.01957}, 2023.

\bibitem{rafailov2024direct}
R.~Rafailov, A.~Sharma, E.~Mitchell, C.~D. Manning, S.~Ermon, and C.~Finn.
\newblock Direct preference optimization: Your language model is secretly a reward model.
\newblock {\em NeurIPS}, 36, 2024.

\bibitem{zou2023universal}
A.~Zou, Z.~Wang, J.~Z. Kolter, and M.~Fredrikson.
\newblock Universal and transferable adversarial attacks on aligned language models, 2023.

\end{thebibliography}

\end{document}